\documentclass[conference]{IEEEtran}
\IEEEoverridecommandlockouts
\usepackage{cite}
\usepackage[colorlinks=true, linkcolor=blue, citecolor=blue, urlcolor=blue]{hyperref}
\usepackage{amsmath,amssymb,amsfonts}
\usepackage{algorithmic}
\usepackage{graphicx}
\usepackage{textcomp}
\usepackage{xcolor}
\usepackage{url}
\usepackage[utf8]{inputenc}
\def\BibTeX{{\rm B\kern-.05em{\sc i\kern-.025em b}\kern-.08em
    T\kern-.1667em\lower.7ex\hbox{E}\kern-.125emX}}
\begin{document}

%\title{GAN-Augmented ResNet-50 for Skin Disease Classification with Explainable AI}

\title{XAI-Driven Skin Disease Classification: Leveraging GANs to Augment ResNet-50 Performance}

\author{
\IEEEauthorblockN{Kim Gerard A. Villanueva}
\IEEEauthorblockA{
Department of Computer Science\\
The University of Texas Permian Basin\\
Odessa, Texas, United States\\
villanueva\_k61778@utpb.edu
}
\and
\IEEEauthorblockN{Priyanka Kumar\thanks{Corresponding author: kumar\_p@utpb.edu}}
\IEEEauthorblockA{
Department of Computer Science\\
The University of Texas Permian Basin\\
Odessa, Texas, United States
}
}

\maketitle

\begin{abstract}
Accurate and timely diagnosis of multi-class skin lesions is hampered by subjective methods, inherent data imbalance in datasets like HAM10000, and the "black box" nature of Deep Learning (DL) models. This study proposes a trustworthy and highly accurate Computer-Aided Diagnosis (CAD) system to overcome these limitations. The approach utilizes Deep Convolutional Generative Adversarial Networks (DCGANs) for per-class data augmentation to resolve the critical class imbalance problem. A fine-tuned ResNet-50 classifier is then trained on the augmented dataset to classify seven skin disease categories. Crucially, LIME and SHAP Explainable AI (XAI) techniques are integrated to provide transparency by confirming that predictions are based on clinically relevant features like irregular morphology. The system achieved a high overall Accuracy of 92.50\% and an Area Under the Curve (AUC) of 98.82\%, successfully outperforming various prior benchmarked architectures. This work successfully validates a verifiable framework that combines high performance with the essential clinical interpretability required for safe diagnostic deployment. Future research should prioritize enhancing discrimination for critical categories, such as Melanoma\_NOS (F$1$-Score: 0.8602).
\end{abstract}


\section{Introduction and Motivation}
Skin diseases and cancer are major public health concern, with cancer being a leading cause of death worldwide and skin cancer ranking as the fifth most prevalent type globally \cite{10455953}. The accurate and early diagnosis of skin lesions is vital for effective treatment and improved patient outcomes, especially given that the highly aggressive forms like Melanoma have a $99\%$ survival rate when detected early \cite{8807441,10880715}. However, traditional diagnostic methods relying on subjective visual inspection, expert judgment, and often expensive and invasive biopsies can be time-consuming, inconsistent, and less accurate, particularly in regions with limited access to specialists \cite{10020302,11210939,11052018,10924047,10455953}.

To overcome these limitations, advanced Computer-Aided Diagnosis (CAD) tools utilizing Deep Learning (DL) have emerged as necessary and effective solutions. Deep Convolutional Neural Networks (DCNNs) have shown outstanding performance in automating medical image analysis, with some models even outperforming highly qualified dermatologists in classification accuracy \cite{8641815,11210939}. This advancement is critical for efficiently and accurately classifying multiple skin lesion categories, moving beyond this simple benign and malignant distinction of traditional methods \cite{8641815}. While dermoscopy is the clinical standard for better lesion visualization, manual diagnosis remains difficult, tedious, subjective, and prone to error, underscoring the necessity of developing an automated CAD system \cite{8807441}.

Despite their success, deep learning models face two challenges in real-world application. First, pathological and dermoscopic datasets often suffer from significant class imbalance, where rare diseases are scarce. This limits diversity and leads to poor model performance on minority classes \cite{10455953,10822517}. Second is lack of interpretability, where models often struggle with real-world image variability, leading to poor generalizability. Crucially, their "black box" nature reduces medical practitioners' understanding
and trust in the system's conclusions \cite{11172177}.

This study addresses these two challenges by proposing a robust and trustworthy automated skin lesion classification system specifically focusing on data augmentation using Generative Adversarial Networks (GANs) to synthesize high quality images for minority classes, overcoming the class imbalance and data diversity; usage of Convolutional Neural Network (CNN) model Resnet50 architecture, a powerful proven CNN framework, for classifying augmented dataset into multiple skin disease categories, and enhancing clinical trust with Explainable AI to provide transparency and make the system clinically reliable highlighting the image features driving the model's predictions. By integrating all of these, the study aims to provide a reliable, efficient, and clinically interpretable CAD tool to significantly improve the accuracy and adoption of automated skin diseases diagnosis.

\section{Literature Survey}

The foundation of modern skin lesion classification research is built upon large, publicly available datasets, primarily the HAM10000 and ISIC datasets, which provide a wide array of dermatoscopic images. Studies focusing on multi-class classification, such as that by \cite{10455953} and \cite{10880715}, predominantly utilized the HAM10000 dataset, which consists of 10,015 images across seven different skin disease classes. The ISIC challenge datasets (ISIC 2017, ISIC 2018, ISIC 2019) were also critical, used by \cite{8641815}, \cite{8807441}, and \cite{10924047} for their seven- and eight-way classification tasks.

The methodological approach predominantly used involves Transfer learning using pre-trained CNN architectures. This technique leverages the feature extraction capabilities of models pre-trained on massive non-dermatological datasets like ImageNet, which is essential for medical image analysis where labeled data is scarce. The most frequently adopted model was Resnet50. \cite{10455953} utilized the ResNet-50 CNN, training it for 150 epochs with the SGD optimizer, achieving a high overall accuracy of 91.71\%. The study by \cite{8807441} employed a cascaded approach where a pre-trained ResNet-50 performed the final classification after a novel Full Resolution Convolutional Network (FrCN) handled the image segmentation task, resulting in 93.8\% binary classification accuracy. The model in the paper by \cite{10533593}, \cite{10918441}, and \cite{10872072} explicitly leveraged transfer learning with a modified ResNet-50 architecture, incorporating global average pooling by adding a layer to efficiently reduce the high-dimensional feature maps (the output of the last convolutional block) into a single, fixed-size feature vector. This improves robustness and reduces the number of parameters, also incorporating dense layers to handle the complex, final decision-making, and tailor it for skin disease classification. Furthermore, \cite{10915335} successfully used a fine-tuned ResNet50 by changing the final fully connected layer to classify three different skin types (normal, oily, dry), demonstrating the architecture's versatility beyond lesion classification.

To overcome the limitation of single models, \cite{8641815} proposed a deep ensemble learning approach, combining ResNet50 and InceptionV3 outputs to achieve superior validation accuracy (89.9\%) and improved classification between confusing lesions. \cite{10880715} explored lightweight models for real-time diagnosis, with the YOLOv8 variants, showing the YOLOv8n-cls model's exceptional speed (0.5 ms inference time).

Generative Adversarial Networks (GANs) are a crucial component across these dermatological classification studies, primarily serving as a powerful tool for data augmentation to overcome the dual challenges of data scarcity and class imbalance inherent in medical image analysis. Researchers consistently leveraged GANs to generate high-quality synthetic images of skin lesions, which effectively expand and diversify the training datasets for CNNs. For instance, the proposed CNN-GAN model by \cite{10689851} specifically utilized GANs to inoculate the approach to circumstances where labelled data is scarce, achieving a notable increase in classification accuracy to 89\% over the traditional CNN baseline of 83\%. Similarly, \cite{11108424} introduced DermaGAN to synthesize diverse images, confirming that this synthetic augmentation strengthens CNN diagnostic skills where annotated data is sparse. Focusing on image quality, \cite{11071055} employed an ESRGAN-based approach to ensure the synthetic images preserved critical diagnostic features, leading to a significant accuracy improvement from 75.34\% on the original dataset to 89.51\% on the augmented one. Furthermore, \cite{11020485} and \cite{10987559} utilized GANs to address specific fairness and generalization issues; \cite{11020485} generated images for rare skin diseases to balance the dataset, while \cite{10987559} used GANs to generate synthetic images of dark skin lesions to mitigate biases against underrepresented skin tones, promoting more equitable dermatological diagnostics. Furthermore, in histopathological contexts, \cite{10822517} developed a GAN to generate high-quality images for minority classes, directly tackling the class imbalance problem and improving data diversity. Therefore, GANs are instrumental in improving the robustness, accuracy, and generalization of deep learning models by providing abundant, high-quality training data.

Explainable AI (XAI) is centrally integrated into dermatological deep learning studies to overcome the critical "black-box" challenge and build clinical trust \cite{11172177}. The primary function of XAI is to interpret the complex decision-making of the AI model and articulate the basis for a given skin disease diagnosis \cite{11022338}. Techniques like LIME are used to visualize the specific image features, such as superpixels or segmentation masks, that directly influence the prediction \cite{11021908}, \cite{11172177}. This visualization allows practitioners to validate the model's reasoning against clinically significant features like lesion borders and asymmetry. By providing this transparency, XAI ensures the model is not only accurate but also dependable and readily accepted in clinical settings \cite{11022338}.  Ultimately, XAI serves as the essential bridge between high-performing deep learning classifiers and the practical requirements of clinical use.

\section{Methodology}
\begin{figure}[htbp]
    \centering
    \includegraphics[width=0.7\linewidth]{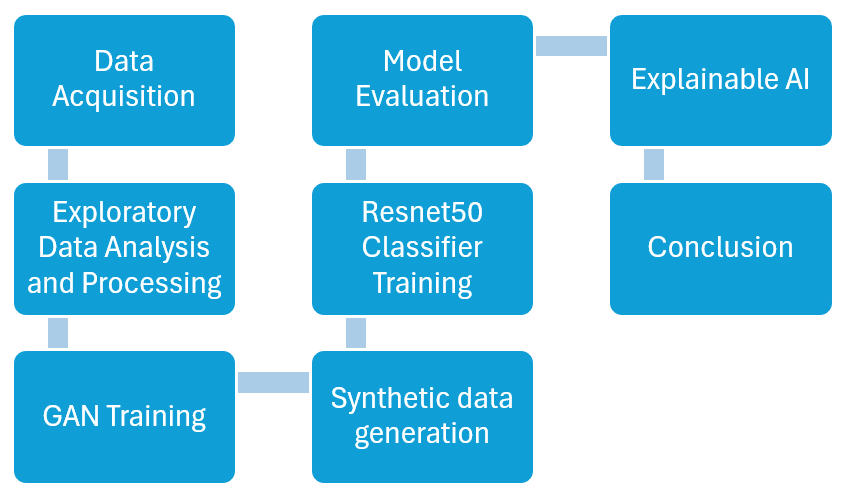}
    \caption{Workflow of the proposed methodology}
    \label{fig:flow}
\end{figure}

The proposed system for skin disease classification using the ResNet50 CNN architecture consists of several stages. First, dermatoscopic images representing seven types of skin lesions are collected from the HAM10000 dataset. Next, the images undergo preprocessing, including cleaning, resizing, normalization, and visualization. Class imbalance is addressed using GAN-based synthetic image generation, and the dataset is stratified into training, validation, and test subsets. The ResNet50 model is then trained on the preprocessed training data to classify the skin lesions, and its performance is evaluated using standard metrics. Finally, explainability techniques are applied to interpret model predictions, and the results are analyzed to draw conclusions regarding the system’s effectiveness. An overview of the proposed system is presented in Fig.~\ref{fig:flow}.

\subsection{Data collection}
The study used an extended version of the HAM10000 dataset from the ISIC Archive \cite{tschandl2018ham10000}, a publicly available multi-source repository of dermatoscopic images representing seven types of pigmented skin lesions: melanocytic nevi, keratosis subtypes, basal cell carcinoma, squamous cell carcinoma, dermatofibroma, and melanoma. The images were originally collected by the ViDIR Group at the Medical University of Vienna and a dermatology practice associated with the University of Queensland, Australia.
While the canonical HAM10000 dataset contains 10,015 images, the version used in this study included 11,540 images with valid metadata. These images were used to create stratified training, validation, and test sets, ensuring balanced representation across lesion classes.

\subsection{Data Pre-processing}
The initial preprocessing phase focused on preparing the image dataset for both the Generative Adversarial Network (GAN) training and the subsequent classifier training, with a primary emphasis on addressing class imbalance.

\begin{figure}[htbp]
    \centering
    \includegraphics[width=1.0\linewidth]{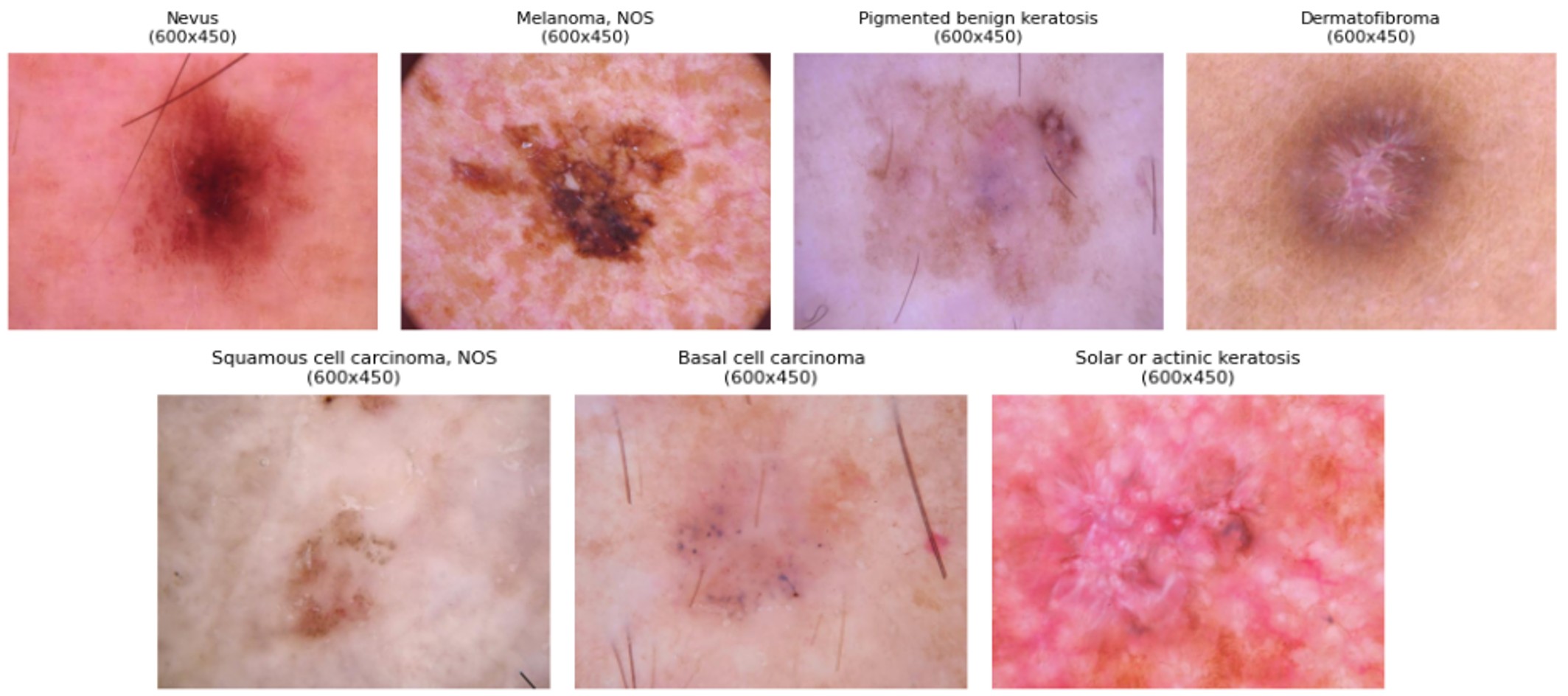}
    \caption{Representative images of skin lesions per category}
    \label{fig:ogdata}
\end{figure}

Per class distribution was analyzed to quantify the data and extent of data imbalance and examine representative images for each class. Fig.~\ref{fig:ogdata} presents a representative image for each skin lesion per category in 600 x 450 pixel size. Table~\ref{tab:distribution} shows the class distribution. Majority class was Nevus and the rest are minority classes with Solar or actinic keratosis being the most minority class. 

\begin{table}[htbp]
    \centering
    \includegraphics[width=0.7\linewidth]{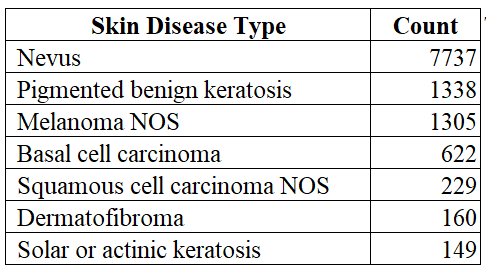} 
    \caption{Skin disease class distribution}
    \label{tab:distribution}
\end{table}

Original images of size 600 × 450 pixels were resized using two pipelines: to 128 × 128 pixels for GAN-based synthetic image generation and to 224 × 224 pixels for ResNet-50 classification. For both resolutions, images were organized into class-specific folders to enable isolated, per-class training. 

\subsection{Synthetic Data Generation}\label{AA}
\begin{figure}[htbp]
    \centering
    \includegraphics[width=1.0\linewidth]{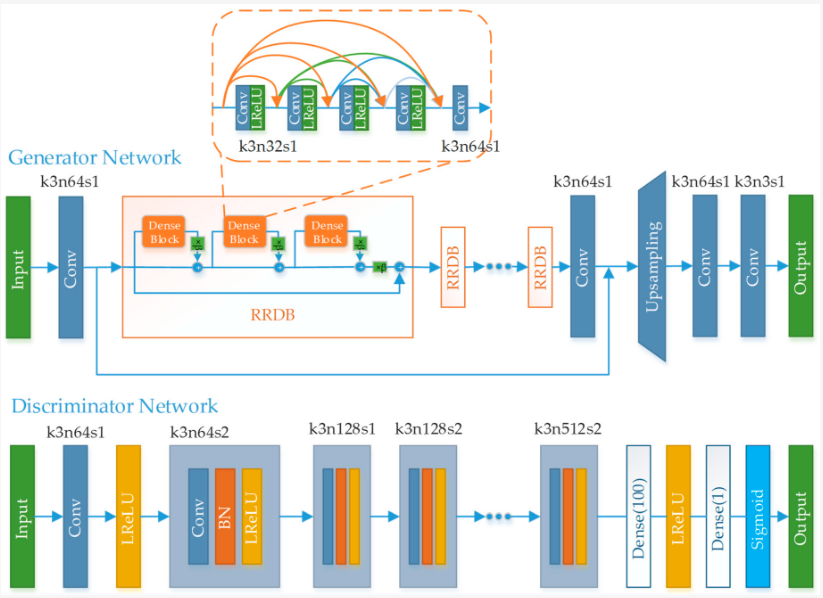}
    \caption{Generative Adversarial Network (GAN) Framework}
    \label{fig:dcgan}
\end{figure}

To address class imbalance, a per-class Deep Convolutional Generative Adversarial Network (DCGAN) strategy was applied exclusively to the minority classes, allowing the model to learn the unique features of underrepresented lesions without interference from the majority class (Nevus). Generative Adversarial Networks (GANs), a prominent framework for generative modeling, utilize a competitive training setup between two neural networks—a Generator and a Discriminator—to produce highly realistic synthetic data \cite{11020485,geeksforgeeksgan}. Fig.~\ref{fig:dcgan} illustrates the competitive training process between the Generator (G) and Discriminator (D). The DCGAN architecture specifically implements both G and D using convolutional layers (transpose convolutions for G and strided convolutions for D).

The DCGAN architecture, specifically suited for image tasks, uses transpose convolutional layers in the Generator for upsampling to create images from a random latent vector and strided convolutional layers in the Discriminator for downsampling to distinguish real samples from generated ones \cite{11020485}.

To stabilize training, Batch Normalization was applied across both networks. The training process began by isolating the $128 \times 128$ images and preparing them via data augmentation (random flips, rotations, color jitter, and affine transformations) and normalization to the $[-1, 1]$ range to match the generator’s Tanh activation function. For optimization, low learning rates ($\mathbf{0.00001}$) for both networks were used to stabilize the competitive training, alongside a batch size of 64 and a latent noise vector size ($n_z$) of 100. Binary cross-entropy loss served as the objective function, and label smoothing (0.9 for real and 0.4 for fake images) was applied to further stabilize convergence.

Upon completion, the trained generator’s weights, encoding the minority class features, were used to produce synthetic images until their counts matched the majority class, effectively balancing the dataset for improved CNN training and generalization. Finally, the generated $128 \times 128$ images were denormalized back to the $[0, 1]$ range and resized to $224 \times 224$ for compatibility with the subsequent CNN classifier, utilizing a ResNet50 architecture.

\subsection{CNN model Resnet-50}
\begin{figure}[htbp]
    \centering
    \includegraphics[width=1.0\linewidth]{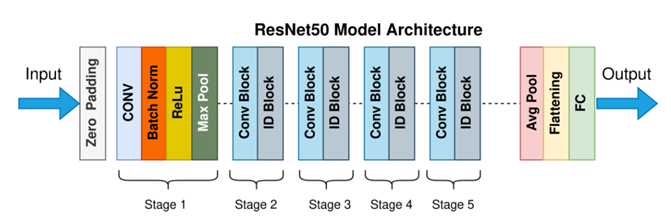}
    \caption{Resnet50 Architecture}
    \label{fig:resnet}
\end{figure}

The classification phase of the research centered on fine-tuning a pre-trained ResNet-50 convolutional neural network  using the GAN-augmented, balanced dataset. Fig.~\ref{fig:resnet} illustrates the structure of the ResNet-50 model, highlighting its use of residual blocks (skip connections) to enable the training of deeper networks. In this study, the model is initialized with ImageNet weights, and its convolutional base is frozen for transfer learning, with only the final classification layers being updated on the augmented skin lesion dataset.

The process begins with data setup, where the images are loaded, split (70\% Train, 15\% Validation, 15\% Test), and pre-processed. Transfer learning was utilized by initializing the base model with ImageNet pre-trained weights, allowing the model to effectively handle the large and diverse dataset. A crucial step here is the application of ImageNet normalization to all data, along with extensive data augmentation (random flip, rotation, color jitter) applied only to the training set.

Model initialization follows the transfer learning approach: the original convolutional layers of ResNet-50 are frozen by setting their parameters to non-trainable, and only a new custom classification head is trained. The model is trained using the Adaptive moment estimation optimizer with a learning rate of 0.00001 to gently update the new layers without altering the frozen weights. Training uses a batch size of 64 and employs early stopping with a patience of 5 epochs based on validation loss to prevent overfitting. The model was trained for 30 epochs. After training, performance is evaluated on the unseen test set.

\section*{Experimental Results and Model Evaluation}
After training, the model’s performance was rigorously evaluated on the unseen test set using standard metrics, including Accuracy, Precision, Recall, F1-Score, and AUC (One-vs-Rest). This evaluation was essential for assessing the model’s effectiveness and its ability to generalize across different skin disease types. During the training process, we monitored both the training and validation accuracy to understand how the model progressed on both seen and unseen data.

\begin{figure}[htbp]
    \centering
    \includegraphics[width=1.0\linewidth]{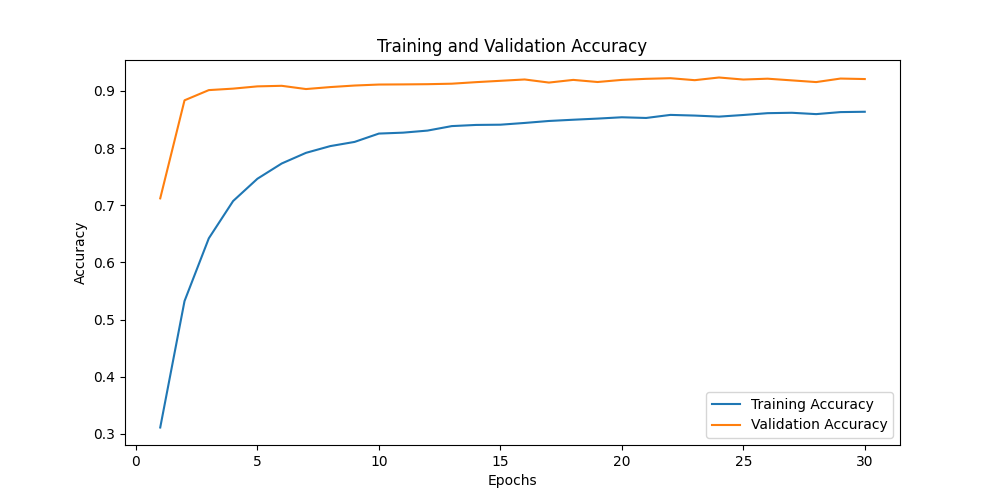}
    \caption{Training and Validation Accuracy Curve}
    \label{fig:accuracycurve}
\end{figure}

Fig.~\ref{fig:accuracycurve} illustrates the model's classification performance on the training data versus the validation data throughout the training epochs. The convergence of these curves confirms the model's stability and strong generalization ability.

The training loss is also monitored, which measures how well the model learned the patterns in the training data, with a decreasing trend indicating better adaptation of the network.

\begin{figure}[htbp]
    \centering
    \includegraphics[width=1.0\linewidth]{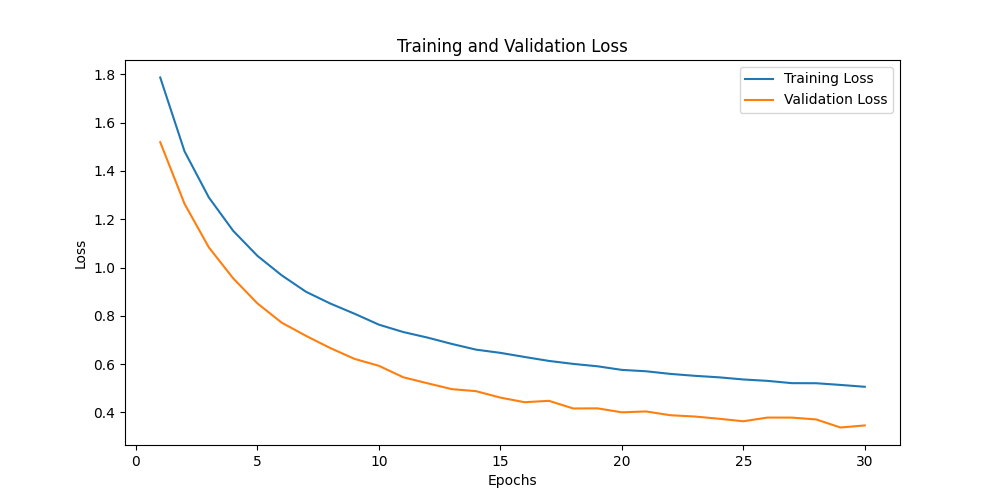}
    \caption{Training and Validation Loss Curve}
    \label{fig:losscurve}
\end{figure}

Fig.~\ref{fig:losscurve}  displays the reduction in the model's loss function value during the training process. The decreasing trend demonstrates the successful learning and adaptation of the ResNet-50 network to the training examples.

The training accuracy reflected how well the model classified the training data, while the validation accuracy served as a crucial indicator of performance on unseen data, confirming that the model did not overfit \cite{10455953}.

The ResNet-50 model demonstrated strong performance, achieving an overall Accuracy of 92.50\% on the test set.

\subsection*{Understanding the Performance Metrics}
The final model assessment achieved the following results presented in Table~\ref{tab:metricsmodel}. To clarify what each metric represents, the following descriptions summarize their purpose and interpretation:

\begin{table}[htbp]
    \centering
    \includegraphics[width=0.5\linewidth]{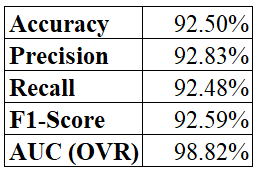} 
    \caption{Model Performance Metrics}
    \label{tab:metricsmodel}
\end{table}

\textbf{Accuracy:} The overall proportion of total predictions (positive and negative) that were correct, determining how close the predicted output is to the actual value \cite{10020302}.

\textbf{Precision:} When the model predicts positive, this specifies how many of those positive values are actually correct \cite{10020302}. This metric is high, indicating the model produces **fewer False Positives** (fewer images of other, non-target diseases are incorrectly classified as the target disease).

\textbf{Recall:} The metric measures how many of the actual positive cases were correctly identified by the model \cite{10020302}. High recall indicates the model produces **fewer False Negatives** (fewer actual cases of the target disease, are incorrectly classified as something else).

\textbf{F$1$-Score:} This score combines Precision and Recall, providing a balanced, overall accuracy measurement of the model. High F$1$-score confirms strong performance across both minimizing false alarms and avoiding missed detections, which is especially important for originally imbalanced multi-class datasets \cite{10020302}.

\textbf{AUC (OVR):} This metric (Area Under the Curve, One-vs-Rest) measures the model's overall ability to correctly distinguish one disease class from all others, averaged across every disease type. The high percentage indicates that the model is excellent at separating all the different skin conditions across all possible decision thresholds, consistently performing much better than random chance \cite{10924047}.

\begin{figure}[htbp]
    \centering
    \includegraphics[width=1.0\linewidth]{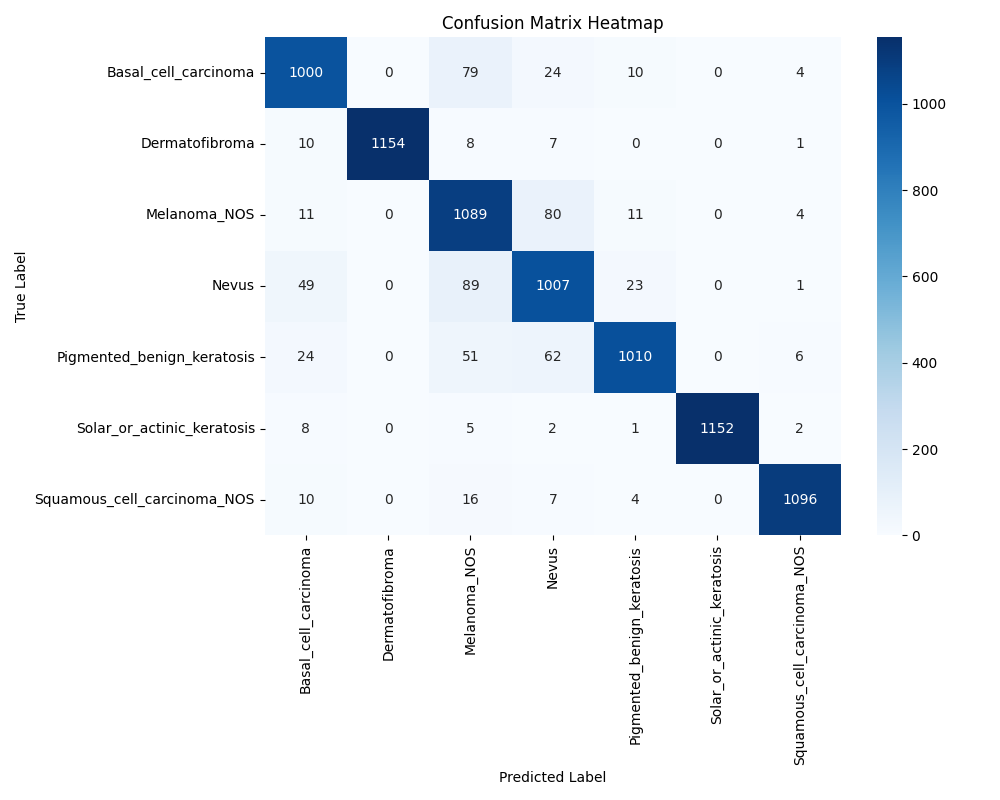}
    \caption{Confusion Matrix}
    \label{fig:matrix}
\end{figure}

These performance metrics offer a comprehensive view of the model's capabilities, calculated for both the entire classification system and each individual disease category. To see the detailed outcomes for each class, including the counts of correct and incorrect predictions, a confusion matrix is presented in Fig.~\ref{fig:matrix}.

\subsection{Class-Wise Performance Analysis} 
The performance metrics were also stratified according to the individual skin lesion categories. The multi-class performance by the GAN-based Resnet50 model was assessed independently across the seven defined classes, with the specific Precision, Recall, and F$1$-Score for each class displayed in Table~\ref{tab:perclass} that presents class-wise performance metrics. 

\begin{table}[htbp]
    \centering
    \includegraphics[width=0.9\linewidth]{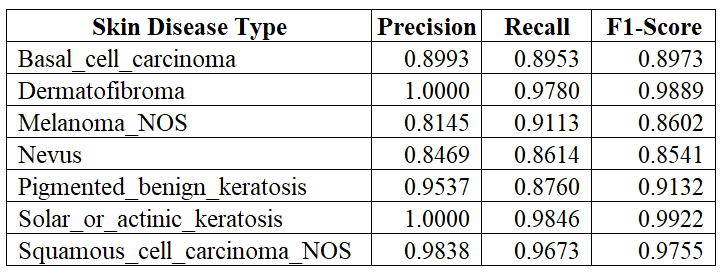} 
    \caption{Performance metrics per class}
    \label{tab:perclass}
\end{table}

The model demonstrated exceptional performance for Solar or actinic keratosis (F$1$-Score 0.9922) and Dermatofibroma (F$1$-Score 0.9889), achieving a perfect Precision of 1.0000 for both, meaning no false positives were recorded for these two diseases. Classification of Squamous\_cell\_carcinoma\_NOS was also very strong (F$1$-Score 0.9755). However, the lowest F$1$-Scores were observed for Nevus (0.8541) and Melanoma\_NOS (0.8602), indicating that these remain the primary classification challenges. Notably, the Melanoma\_NOS class exhibited a significant trade-off, showing high Recall (0.9113)—minimizing false negatives by successfully identifying most actual cases—but lower Precision (0.8145). These result highlight the need for future efforts to improve discrimination among these more challenging skin diseases conditions.

\subsection{Comparison with Previous Studies}
The findings from the present study were compared against prior works that utilized the same dataset. Using the comparison framework presented in \cite{10455953}, and incorporating the results from the three most recent studies \cite{10455953,10020302,Chaturvedi2020}, this study’s performance metrics were added to provide a comprehensive benchmark.

The comparison Table~\ref{tab:comparison} effectively benchmarks the performance of the proposed ResNet-50 model against various deep learning architectures reported in prior studies. The data clearly show that the ResNet-50 model used in this study achieved the highest overall performance, leading all compared models across all four metrics: Accuracy (92.50\%), Precision (92.83\%), Recall (92.59\%), and F$1$-Score (92.59\%). For instance, while other architectures such as NASNetLarge and InceptionV3 achieved high accuracy (up to 91\%) and a Resnet-50 with 91.71\% accuracy, the current ResNet-50 model surpassed them with better balanced metrics. This performance is particularly notable when compared to Study 2, where a standard ResNet-50 achieved significantly lower results (82\% Accuracy), highlighting that the methodology used here—specifically the GAN-augmented dataset and optimized hyperparameter settings (Adam optimizer, 30 epochs, batch size of 64, and a low learning rate of 0.00001)—was key to unlocking the model's superior performance and generalization ability. The table thus validates that the approach taken in this research is highly competitive and effective, achieving state-of-the-art results with the ResNet-50 architecture in skin disease classification.

\begin{table}[htbp]
    \centering
    \includegraphics[width=1.0\linewidth]{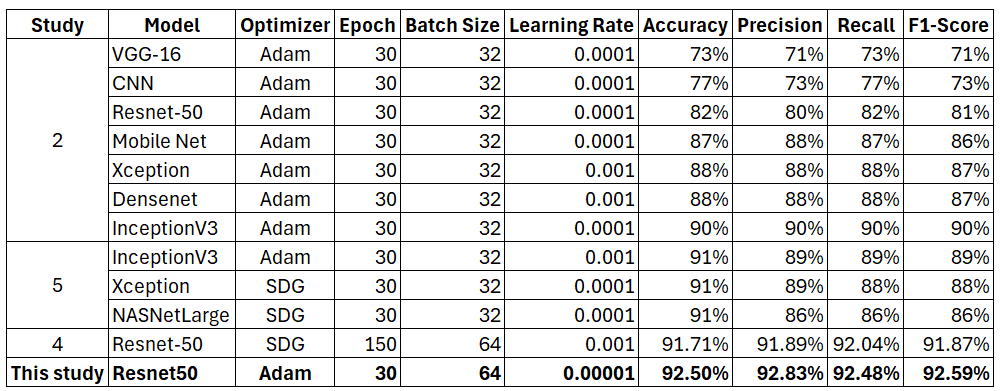}
    \caption{Comparison with previous studies}
    \label{tab:comparison}
\end{table}

\section{Integration of XAI}
Two Explainable AI tools- LIME (Local Interpretable Model-agnostic Explanations) and SHAP (Shapley Additive exPlanations), were integrated with Resnet50 model to make its decision-making process transparent and understandable \cite{11022338}. These methods are essential for moving beyond a "black-box" approach by explicitly identifying which features within an image influence the model's final decision. LIME is used to explain individual predictions by slightly manipulating the input image, observing the resulting output, and then using a simplified model to figure out what ResNet50 was actually focusing on. In contrast, SHAP takes a more comprehensive, mathematically grounded approach. It determines the precise influence of every single pixel on the final result, much like assigning credit in a collaborative effort. By using SHAP as a model-specific tool in our research, we achieved a fairer, overall understanding of the ResNet50 model's logic. Together, the visualizations generated by LIME and SHAP offer a comprehensive, actionable understanding of the ResNet50 model's predictive mechanism, validating its logic and revealing potential biases.

\subsection{Visual Validation using LIME and SHAP}

To validate the model's predictive focus, LIME and SHAP were applied to a representative image classified as Melanoma\_NOS.
 
\begin{figure}[htbp]
    \centering
    \includegraphics[width=1.0\linewidth]{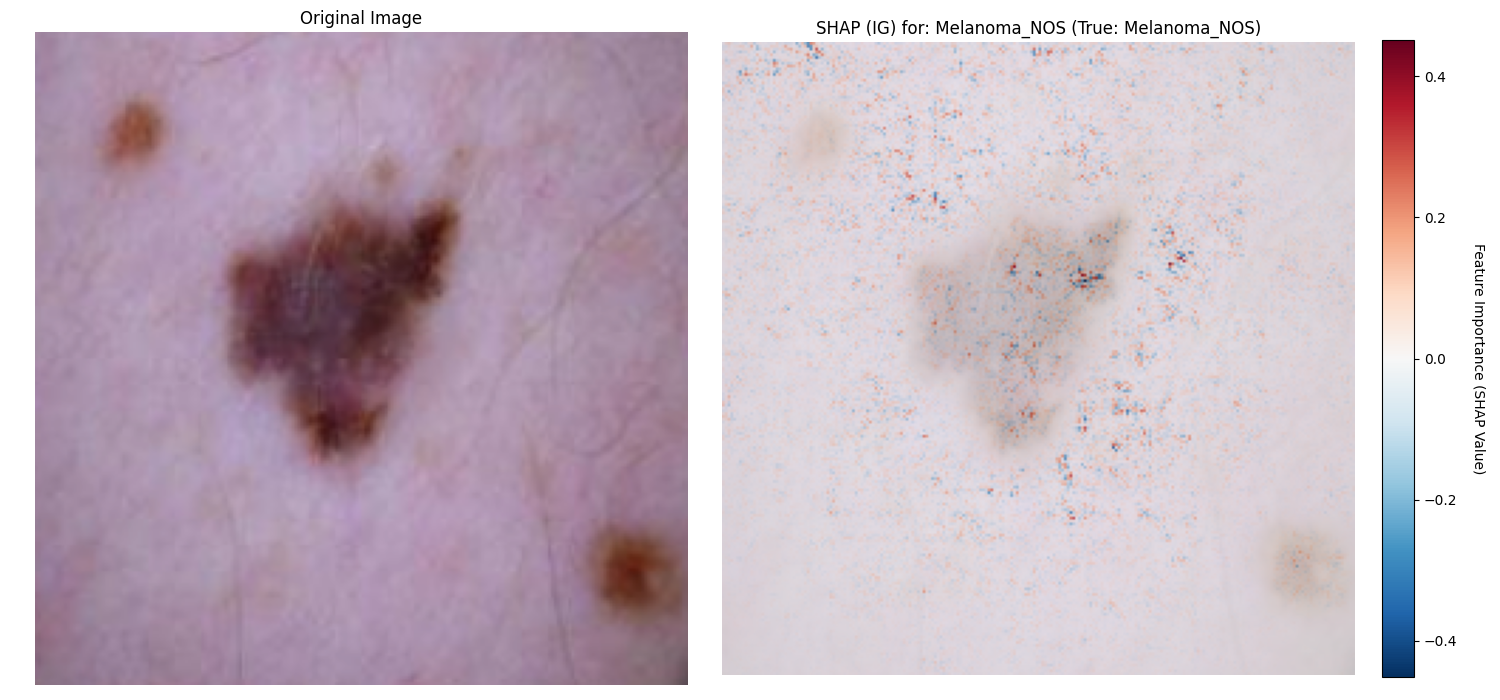}
    \caption{SHAP visualization}
    \label{fig:shap}
\end{figure}

The SHAP visualization Fig.~\ref{fig:shap}  provides a detailed, pixel-level explanation for the "Melanoma\_NOS" classification using a heat map. The feature importance for every pixel is clearly indicated by color by using a color scale where Red indicates a positive influence (features supporting the prediction) and Blue indicates a negative influence (features opposing the prediction) \cite{11022338}. Dark red pixels are densely concentrated over the dark, irregular core of the lesion, demonstrating that features such as the irregular shape, uneven color, and high contrast/texture are the most important features positively driving the model toward the melanoma prediction. Conversely, the blue regions and light tints on the surrounding skin indicate features (like normal skin color) that slightly oppose the prediction, but their influence is overwhelmed by the strong positive signal from the pathological core, ultimately leading to the correct and confident final classification.

\begin{figure}[htbp]
    \centering
    \includegraphics[width=0.6\linewidth]{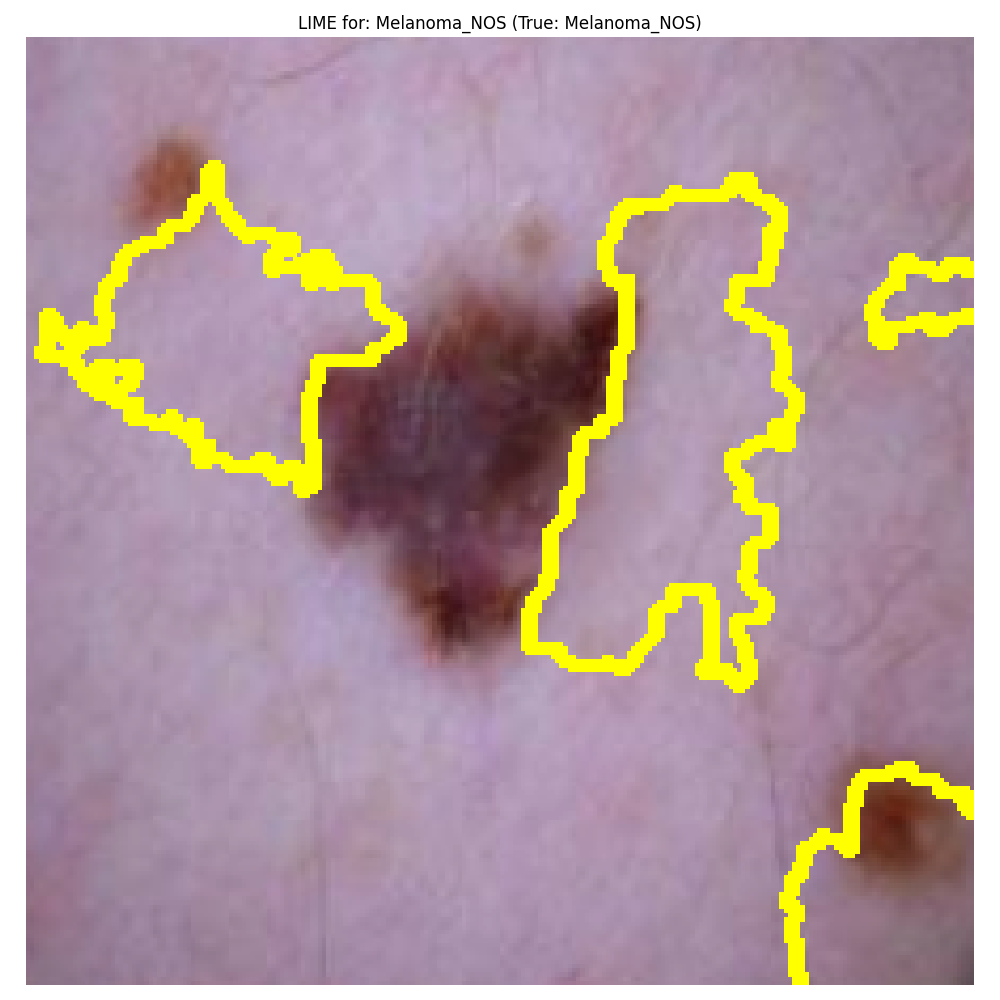}
    \caption{Lime-based visualization of prediction explanations.}
    \label{fig:lime}
\end{figure}

The LIME visualization Fig.~\ref{fig:lime} serves as a local, explanation of the ResNet50 model's prediction for the Melanoma\_NOS lesion. The key output is the prominent yellow outline superimposed on the original image, which directly identifies the region of the lesion that the model deemed most influential. This boundary encompasses the central, darkest, and most irregular area of the skin lesion. LIME determined that this region's features—particularly its complex border and dense pigmentation—were the primary factors used by the simplified, interpretable model to approximate the complex ResNet50's decision, thus confirming that the deep learning model focused correctly on the clinically significant features of the lesion.

\section{Conclusion and Future Work}
The proposed system successfully delivers a reliable, efficient, and clinically interpretable Computer-Aided Diagnosis (CAD) tool for multi-class skin lesion classification by integrating GAN-based data augmentation and a fine-tuned ResNet-50 classifier. By leveraging Generative Adversarial Networks (GANs) to address the critical issue of class imbalance within the HAM10000 dataset, the approach ensures that minority classes are adequately represented, leading to robust and generalized model learning. The resulting model demonstrated strong quantitative performance, achieving an overall Accuracy of 92.50\% and a high Macro-AUC of 98.82\%, thus outperforming various deep learning architectures benchmarked in prior studies. Furthermore, the integration of Explainable AI (XAI) techniques, LIME and SHAP, provides essential transparency by visually confirming that the model's predictions are correctly based on clinically significant features, such as the irregular morphology and dense pigmentation of lesions, thereby fostering the necessary trust for clinical deployment. While the system shows excellent performance, particularly in classes like Solar Keratosis, future efforts must focus on improving the classification boundaries for highly challenging categories like Melanoma\_NOS, which exhibited the lowest F$1$-Score (0.8602), to further minimize the risk of dangerous False Negative predictions. Ultimately, this work establishes a verifiable and highly accurate framework that successfully bridges the gap between state-of-the-art deep learning performance and the requirement for clinical interpretability in dermatological diagnostics.

\section{Acknowledgments}
I sincerely thank my professor, Dr. Priyanka Kumar, for her guidance and support on this project. Her expertise was essential to the success of this study. I am deeply grateful for her mentorship.

\section{Funding}
This work was not supported by any funding agency or grant.

\bibliographystyle{IEEEtran}
\bibliography{references}

\end{document}